# An Algorithm to Extract Rules from Artificial Neural Networks for Medical Diagnosis Problems

S. M. Kamruzzaman[1] and Md. Monirul Islam[2]

[1]Department of Information and Communication Engineering, University of Rajshshi, Rajshahi-6205, Bangladesh
smzaman@gmail.com

[2]Department of Computer Science and Engineering, Bangladesh University of Engineering and Technology, Dhaka-1000, Bangladesh
mdmonirulislam@cse.buet.ac.bd

## Abstract

Artificial neural networks (ANNs) have been successfully applied to solve a variety of classification and function approximation problems. Although ANNs can generally predict better than decision trees for pattern classification problems, ANNs are often regarded as black boxes since their predictions cannot be explained clearly like those of decision trees. This paper presents a new algorithm, called rule extraction from ANNs (REANN), to extract rules from trained ANNs for medical diagnosis problems. A standard three-layer feedforward ANN with four-phase training is the basis of the proposed algorithm. In the first phase, the number of hidden nodes in ANNs is determined automatically by a constructive algorithm. In the second phase, irrelevant connections and input nodes are removed from trained ANNs without sacrificing the predictive accuracy of ANNs. The continuous activation values of the hidden nodes are discretized by using an efficient heuristic clustering algorithm in the third phase. Finally, rules are extracted from compact ANNs by examining the discretized activation values of the hidden nodes. Extensive experimental studies on three benchmark classification problems, i.e. breast cancer, diabetes and lenses, demonstrate that REANN can generate high quality rules from ANNs, which are comparable with other methods in terms of number of rules, average number of conditions for a rule, and predictive accuracy.

## I. Introduction

The last two decades have seen a growing number of researchers and practitioners applying artificial neural networks (ANNs) for pattern classifications and function approximations [3], [13], [20]. While the predictive accuracy of ANNs is often higher than that of other methods or human experts, it is generally difficult to understand how ANNs arrive at a particular conclusion due to the complexity of ANN architectures. Even an ANN with single hidden layer, it is generally impossible to explain why a certain pattern is classified as a member of one class and another pattern as a member of another class [10]. It is therefore desirable to have a set of rules to explain how ANNs solve a given problem. This is because the functionality of ANNs represented by a set of rules will be more comprehensible to human users than a set of connection weights of ANNs [8].

There are a number of works in the literature to explain the functionality of ANNs by extracting rules from trained ANNs [1], [2]. The main problem of existing work is that they determine the number of hidden neurons in ANNs manually. Thus the prediction accuracy and rules extracted from trained ANNs may not be optimal since the performance of ANNs is greatly dependent on their architectures. Furthermore, rules extracted by existing algorithms are not simple as a result it is difficult to understand for users.

This paper proposes a new algorithm, called rule extraction from ANNs (REANN), to extract rules from trained ANNs for medical diagnosis problems. A standard three-layer feedforward ANN with four-phase training is the basis of REANN. The salient feature of REANN is that it does not require many user specified parameters for extracting rules. In addition, an efficient clustering algorithm is used in REANN to discretize the continuous values of hidden nodes so that rules can be extracted easily by using discretized values.

The rest of this paper is organized as follows. Section II discusses some related works for extracting rules from trained ANNs. Section III describes our REANN algorithm in details. Section IV presents results of our experimental study. Finally, Section V concludes the paper with a brief summary and a few remarks.

## II. Related Work

A number of algorithms have been developed for extracting rules from trained ANNs in the last two decades [7]-[13]. In this section, we describe some algorithms that are related to the present work. The problems of existing algorithms are also described in this section.

Two methods for extracting rules from ANN are described by Towell and Shavlik [4]. The first method is the subset algorithm, which searches for subsets of connections to a unit whose summed weight exceeds the bias of that node [5]. The major problem with the subset algorithm is that the cost of finding all subsets increases as the size of ANNs increases. The second method i.e. MofN is an improvement of the subset method that is designed to explicitly search for M-of-N rules from knowledge based ANNs [6]. It checks a group of connections instead of a single connection in ANNs to find their contribution in node's activation. This is done by clustering the connections of ANNs. The problems of MofN are it uses threshold activation function, which is not continuous and uses fixed number of hidden nodes that require prior knowledge of the problem to be solved.

In 1995, H. Liu and S. T. Tan [7] propose, a simple and fast algorithm X2R that can be applied to both numeric and discrete data for generating rules. X2R can generate concise rules from raw data sets by using first order information. It can generate perfect rules in the sense that the error rate of the rules is not worse than the inconsistency rate found in the original data. The problem of X2R is that rules generated by it are order sensitive i.e. generated rules should be fired in sequence.

R. Setiono and H. Liu [8] present a novel way to understand ANNs by extracting rules with a three phase algorithm. A weight decay backpropagation network is built in the first phase so that important connections are reflected by their bigger weights. In the second phase, the network is pruned in such a way so that insignificant connections are deleted while its predictive accuracy is still maintained. In the third phase, rules are extracted by recursively discretizing the hidden unit activation values. The problem of three phase algorithm is that the discretizing algorithm used to discretize the output values of hidden nodes is not efficient.

In 2002, R. Setiono et al. [13] proposed a new method REFANN (rule extraction from function approximating neural networks) for extracting rules from trained ANNs for nonlinear regression. It is shown REFANN can produce rules that are almost as accurate as the original ANNs from whom rules are extracted. For some problems, REFANN extracts few rules that represent useful knowledge for explaining problems easily. REFANN approximates the nonlinear hyperbolic tangent activation function of the hidden nodes by using a simple three-piece or five-piece linear function. It then generates rules in the form of linear equations from trained ANNs. The problem of REFANN is that it needs to divide the continuous hidden node activation into three-piece or five-piece linear function, which may not be possible for complex problems.

The problems of existing algorithms are summarized as follows:
  (i)    Use predefined and fixed number of hidden nodes that require human experience and prior knowledge of the problem to be solved,
  (ii)   Clustering algorithms used to discretize the output values of hidden nodes are not efficient,
  (iii)  Computationally expensive, and
  (iv)   Could not produce concise rules.

## III.   The REANN Algorithm

The aim of this section is to introduce rule extraction algorithm REANN for understanding how an ANN solves a given problem. Although REANN is applied in medical domain in this work, it can be applied to other domain also. The aim of REANN is to search for simple rules with high predictive accuracy.

In comparison with other existing algorithms in the literature, the major advantages of REANN include: (i) it can determine near optimal ANN architectures automatically by using a constructive-pruning strategy; (ii) it uses an efficient method to discretize the output values of hidden nodes; (iii) it is computationally inexpensive; and (iv) it can extract rules that are concise, comprehensible, order insensitive and highly accurate.

The major steps of REANN are summarized in Fig. 1 and explained further as follows:

**Step 1** Create an initial ANN architecture. The initial architecture has three layers i.e. an input, an output and a hidden layer. The number of nodes in the input and output layers is the same the number of inputs and outputs of the problem, respectively. Randomly initialize all connection weights of the ANN within a small range.

**Step 2** Determine the number of hidden nodes in the ANN by using a basic constructive algorithm.

**Step 3** Remove the redundant input nodes and connections by using a basic pruning algorithm. When pruning is completed, the ANN architecture contains only important nodes and connections.

**Step 4** Discretize the outputs of hidden nodes by using an efficient heuristic clustering algorithm. The reason for discretization is that the outputs of hidden nodes are continuous therefore rules cannot be easily extracted from the ANN.

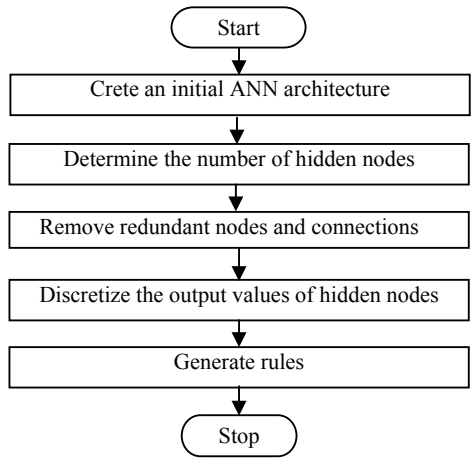

**Fig. 1** Flow chart of the REANN algorithm.

**Step 5** Generate rules that map the inputs and outputs relationships. The task of rules generation is accomplished in three steps. In the first step, rules are generated by using rule extraction algorithm REx to describe the outputs of the ANN in terms of the discretized output values of its hidden nodes. In the second phase, rules are generated by REx to describe the discretized output values of hidden nodes in terms of their inputs. Finally, rules are generated by combining the rules generated in the first and second steps.

It is seen that REANN is very straightforward. However, REANN is consisted of four phases, which are implemented sequentially one by one. In the following subsections, each phase is described elaborately and the reasons for utilizing different techniques in each phase are also explained.

## A. Constructive Algorithm

One drawback of the traditional backpropagation algorithm is the need to determine the number of nodes in the hidden layer prior to training [14]-[17]. REANN uses a basic constructive algorithm based on dynamic node creation algorithm proposed by T. Ash [18].

The major steps of the constructive algorithm used in REANN are summarized in Fig. 2 and explained further as follows:

**Step 1** Create an initial ANN consisting of three layers, i.e., an input, an output, and a hidden layer. The number of nodes in the input and output layers is the same as the number of inputs and outputs of the problem. Initially the hidden layer contains only one node. Randomly initialize all connection weights within a certain range.

**Step 2** Train the network on the training set by using BP algorithm until the error is almost constant for a certain number of training epochs τ that is specified by the user.

**Step 3** Compute the error of the ANN based on the validation set. If the error is found unacceptable (i.e., too large), then assume that the ANN has inappropriate architecture, and go to the next step. Otherwise, stop the training process. The error $E$ is calculated according to the following equations.

$$E(w,v) = \frac{1}{2}\sum_{i=1}^{k}\sum_{p=1}^{C}(S_{pi} - t_{pi})^2 \quad (1)$$

where k is the number of patterns and C is the number of output nodes. $t_{pi}$ and $S_{pi}$ are the target and actual outputs for the ith pattern of the pth output node. The actual output $S_{pi}$ is calculated according to the following equation.

$$S_{pi} = \sigma(\sum_{m=1}^{h}\delta((x_i)^T w_m)v_{pm}) \quad (2)$$

Here $h$ is the number of hidden nodes in the network, $x_i$ is an n-dimensional input pattern, $i=1, 2,..., k$, $w_m$ is an p-dimensional vector weights for the arcs connecting the input layer and the m-th hidden node, $m=1, 2, ..., h$, $v_m$ is a c-dimensional vector of weights for the arcs connecting the m-th hidden node and the output layer. The activation function for the output layer is sigmoid function $\sigma(y) = 1/(1+e^{-y})$ and for the hidden layer hyperbolic tangent function $\delta(y) = (e^y - e^{-y})/(e^y + e^{-y})$.

**Step 4** Add one hidden node to the hidden layer. Randomly initialize the weights of the newly added node and go to step 2.

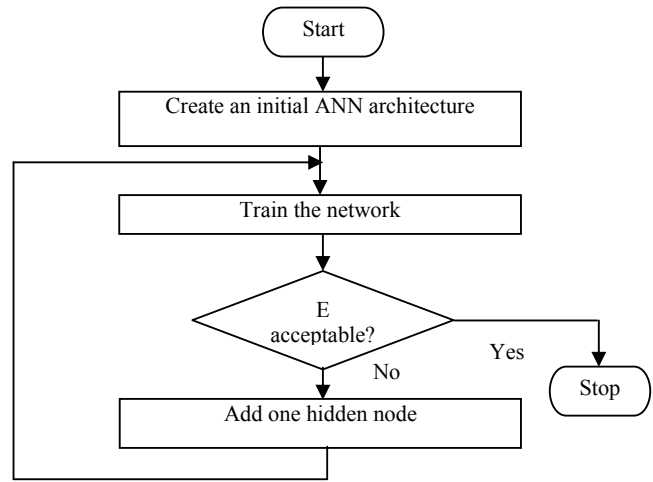

**Fig. 2** Flow chart of the constructive algorithm used in REANN.

Although other architecture determination algorithms, such as pruning [24] and evolutionary algorithms [23], could be used in REANN, the reasons for using constructive algorithm are four folds. Firstly, it is straightforward in constructive algorithms to specify an initial network, while it is problematic in pruning algorithms, one does not know in practice how big the initial network should be. Secondly, constructive algorithms always search for small network solutions first. They are thus computationally more efficient than pruning algorithms, in which the majority of the training time is spent on networks larger than necessary. Because of smaller solutions, the ANN is less likely to overfit the training data and, thus, more likely to generalize better. Thirdly, the strong convergence of a constructive algorithms follows directly from its universal approximation ability. Finally, a constructive approach usually requires a relatively small number of user specified parameters. The use of many user specified parameters requires a user to know rich prior knowledge, which often does not exist for complex real world problems.

## B. Pruning Algorithm

Pruning techniques begin by training a larger than necessary network and then eliminate the weights and nodes that are deemed redundant [21], [22]. Since the nodes in the hidden layer are determined automatically in constructive fashion in REANN, the aim of pruning algorithm is to remove unnecessary connections and input nodes from the ANN obtained by the constructive algorithm. Typically, methods for removing connections from ANNs involve adding a penalty term to the error function. It is hoped that by adding a penalty term to the error function, unnecessary connections will have small weights, and therefore pruning can reduce the complexity of the ANN significantly. The simplest and most commonly used penalty term is the sum of the squared weights [19].

The pruning algorithm used in REANN is briefly described below. This pruning algorithm removes the connections of the ANN according to the magnitudes of their weights. Since the eventual goal of REANN is to get a set of simple rules that describe the classification process, it is important that all unnecessary connections and nodes must be removed. In order to remove as many connections as possible, the weights of the network must be prevented from taking values that are too large. At the same time, weights of irrelevant connections should be encouraged to converge to zero. The penalty function is found to be particularly suitable for these purposes [19].

The steps of the weight-pruning algorithm are summarized in Fig. 3 and explained further as follows:

**Step 1** Train the network to meet a prespecified accuracy level with the following condition satisfied by all correctly classified input patterns:

$$\max_p |e_{pi}| = \max_p |S_{pi} - t_{pi}| \leq \eta_1, \; p = 1, 2, ..., C. \quad (3)$$

Let $\eta_1$ and $\eta_2$ be positive scalars such that $\eta_1 + \eta_2 < 0.5$ ($\eta_1$ is the error tolerance, $\eta_2$ is a threshold that determines if a weight can be removed), where $\eta_1 \in [0, 0.5)$. Let (w, v) be the weights of this network.

**Step 2** Remove the connections between the input nodes and the hidden nodes and between the hidden nodes and the output nodes. This task is accomplished in two phases. In the first phase, connections between the input nodes and hidden nodes are removed.

For each $w_{ml}$ in the network, if $\max_p |v_{pm} w_{ml}| \leq 4\eta_2,$ (4)

then remove $w_{ml}$ from the network.

In the second phase, connections between the hidden nodes and output nodes are removed.

For each $v_{pm}$ in the network, if $|v_{pm}| \leq 4\eta_2,$ (5)

then remove $v_{pm}$ from the network.

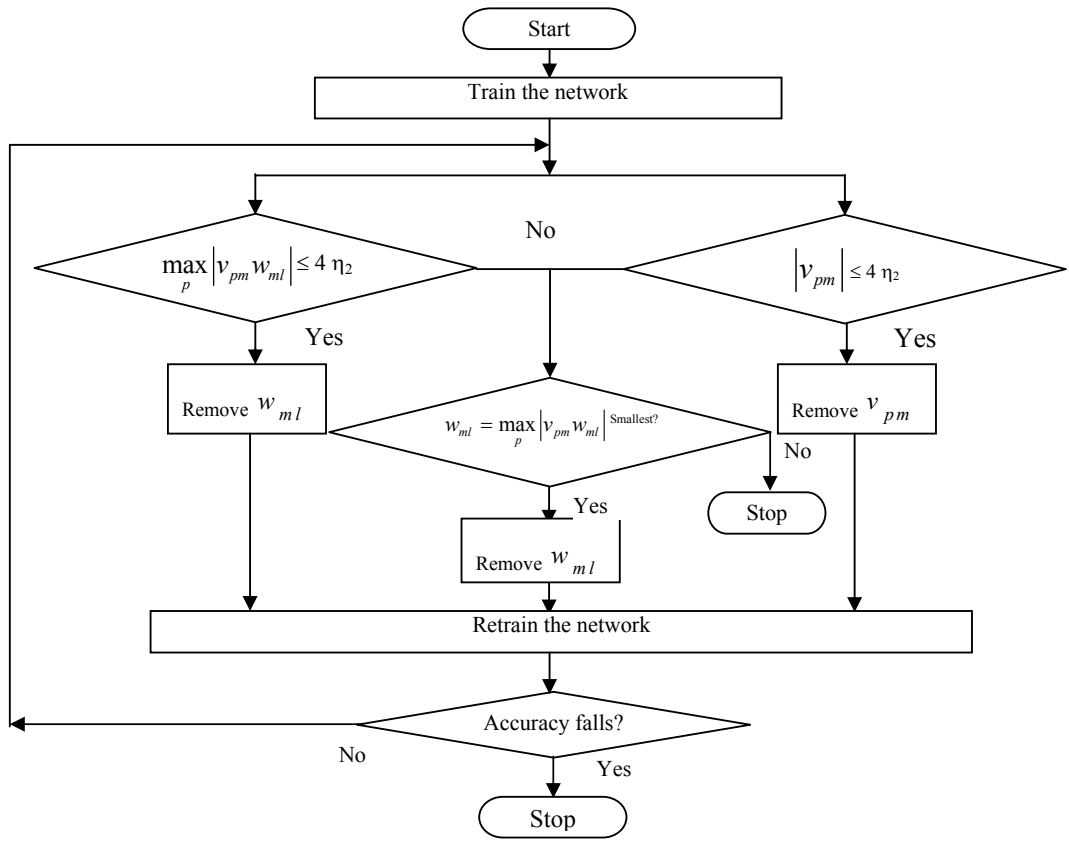

**Fig. 3** Flowchart of the pruning algorithm.

**Step 3** Remove the connections between the input nodes and the hidden nodes further. If no weight satisfies condition (4) or (5), then for each $w_{ml}$ in the network, compute $w_{ml} = \max_{p} |v_{pm} w_{ml}|$ and remove the smallest $w_{ml}$.

**Step 4** Retrain the network and calculate the classification accuracy of the network.

**Step 5** If the classification rate of the network falls below an acceptable level, then stop and use the previous setting of network weights. Otherwise, go to Step 2.

The pruning algorithm is used in REANN also to reduce the amount of training time. Although it can no longer be guaranteed that the resultant pruned ANN will give the same accuracy rate as the original ANN, the experiments show that many weights can be eliminated simultaneously without deteriorating the performance of the ANN. The two conditions (4) and (5) for pruning depends on the weights for connections between the input and hidden nodes and between the hidden and output nodes. It is imperative that during the training, these weights be prevented from getting too large. At the same time, small weights should be encouraged to decay rapidly to zero.

## C. Heuristic Clustering Algorithm

The process of grouping a set of physical or abstract objects into classes of similar objects is called clustering. A cluster is a collection of data or objects that are similar within the same cluster and dissimilar to data or objects in other clusters [25]. A large number of clustering algorithms exist in the literature including k-means and k-medoids [26], [27]. The choice of clustering algorithm depends both on the type of data available and on the particular purpose and application.

After applying pruning algorithm in REANN, the ANN architecture produced by constructive algorithm contains only important connections and nodes. Nevertheless, rules are not readily extractable because the hidden node activation values are continuous. The discretization of these values paves the way for rule extraction.

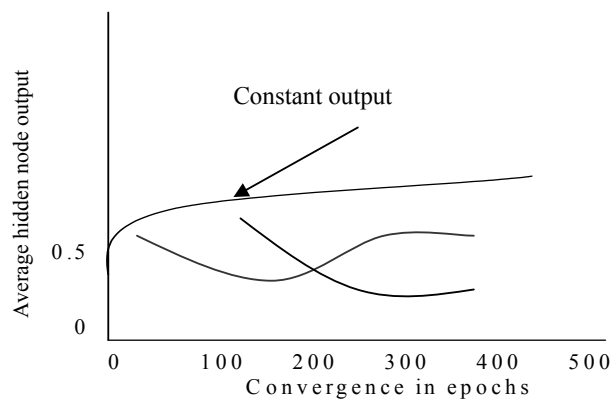

**Fig. 4** Output of hidden nodes.

It is found that some hidden nodes of an ANN maintain almost constant output while other nodes change continuously during the whole training process [28]. Fig. 4 shows typical example where one hidden node maintains almost constant output after some training epochs and the outputs of two hidden nodes are continuously changing.

In REANN, no clustering algorithm is used when hidden nodes maintain almost constant output. If the outputs of hidden nodes do not maintain constant value, a heuristic clustering algorithm is used to discretize the output values of hidden nodes.

The steps of the heuristic clustering algorithm are summarized in Fig. 5 and are explained further as follows:

**Step 1** Let $\varepsilon \in (0, 1)$. D is the activation values in the hidden node. $\delta_1$ is the activation value for the first pattern.

The first cluster, $H(1) = \delta_1$, count = 1, and sum $(1) = \delta_1$, set D = 1.

**Step 2** For each pattern $p_i$, checks whether subsequent activation values can be clustered into one of the existing clusters. The distance between an activation value under consideration and its nearest cluster, $|\delta - H(\overline{j})|$, is computed. If this distance is less than $\varepsilon$, then the activation value is clustered in cluster $\overline{j}$. Otherwise, this activation value forms a new cluster. Let $\delta$ be its activation value. If there exists an index $\overline{j}$ such that

$$|\delta - H(\overline{j})| = \min_{j\varepsilon\{1,2,......D\}} |\delta - H(j)| \text{ and } |\delta - H(\overline{j})| \leq \varepsilon$$

then set count($\overline{j}$) := count($\overline{j}$)+1, sum($\overline{j}$) := sum($\overline{j}$)+ $\delta$, else D = D+1
$H(D) = \delta$, count $(D) = 1$, sum $(D) = \delta$.

**Step 3** Replace H by the average of all activation values that have been clustered into this cluster: $H(j)$ := sum $(j)$/count $(j)$, j=1, 2, 3, …, D.

**Step 4** Once the activation values of all hidden nodes have been obtained, the accuracy of the network is checked with the activation values at the hidden nodes replaced by their discretized values. An activation value $\delta$ is replaced by $H(\overline{j})$, where index $\overline{j}$ is chosen such that $\overline{j} = \arg\min_j |\delta - H(j)|$. If the accuracy of the network falls below the permitted limit then $\varepsilon$ must be decreased and the clustering algorithm is run again, otherwise stop.

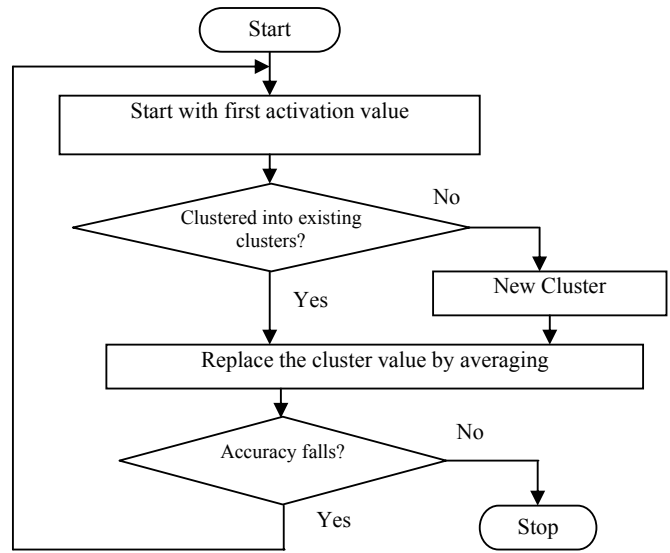

**Fig. 5** Flow chart of the heuristic clustering algorithm.

For a sufficiently small ε, it is always possible to maintain the accuracy of the network with continuous activation values, although the resulting number of different discrete activations can be impractically large. The best ε value is one that gives a high accuracy rate after the clustering and at the same time generates as few clusters as possible. A simple way of obtaining an optimal value for ε is by searching in the interval (0, 1). The number of clusters and the accuracy of the network can be checked for all values of ε = iζ, i= 1, 2… where ζ is a small positive scalar, e.g. 0.10. Note also that it is not necessary to fix the value of ε equal for all hidden nodes.

### D. Rule Extraction Algorithm (REx)

Classification rules are sought in many areas from automatic knowledge acquisition [28], [30] to data mining [31], [32]. They should be explicit, understandable and verifiable by domain experts, and could be modified, extended and passed on as modular knowledge. The REx algorithm described in this section possesses the above mentioned quality and is composed of three major functions:

(i) Rule Extraction- This function iteratively generates shortest rules and remove/marks the patterns covered by each rule until all patterns are covered by the rules;
(ii) Rule Clustering- Rules are clustered in terms of their class levels; and
(iii) Rule Pruning- Redundant or more specific rules in each cluster are removed.

A default rule should be chosen to accommodate possible unclassifiable patterns. If rules are clustered, the choice of the default rule is based on clusters of rules.

The steps of the Rule Extraction (REx) algorithm are summarized in Fig. 6 and explained further as follows:

**Step 1** Extract Rule:
i=0; while (data is NOT empty/marked){
generate $R_i$ to cover the current pattern and differentiate it from patterns in other categories;
remove/mark all patterns covered by $R_i$ ; i++}
The core of this step is a greedy algorithm that finds the shortest rule based on the first order information that can differentiate the pattern under consideration from the patterns of other classes. It then iteratively generates rules and removes the patterns covered by the rules.

**Step 2** Cluster Rule:
Cluster rules according to their class levels. Rules generated in Step 1 are grouped in terms of their class levels. In each rule cluster, redundant rules are eliminated; specific rules are replaced by more general rules.

**Step 3** Prune Rule:
replace specific rules with more general ones;
remove noisy rules;
eliminate redundant rules.

**Step 4** Check whether all patterns are covered by any rules. If yes then stop, otherwise continue.

**Step 5** Determine a default rule:
A default rule is chosen when no rule can be applied to a pattern.

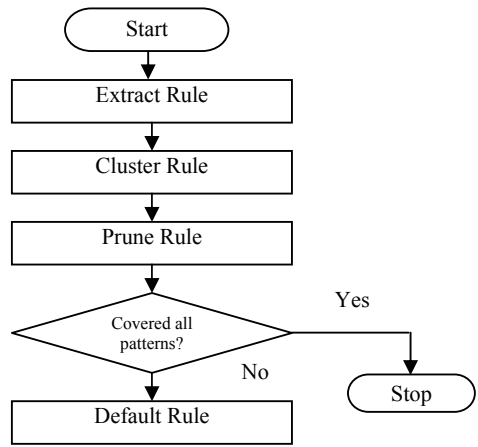

**Fig. 6** Flow chart of the rule extraction (REx) algorithm.

REx exploits the first order information in the data and finds shortest sufficient conditions for a rule of a class that can differentiate it from patterns of other classes. It can generate concise and perfect rules in the sense that the error rate of the rules is not worse than the inconsistency rate found in the original data. The novelty of REx is that the rule generated by it is order insensitive, i.e. rules need not be required to fire sequentially.

## IV. Experimental Studies

This section evaluates the performance of REANN on three well-known benchmark classification problems. These are breast cancer, diabetes and lenses, which are widely used in machine learning and ANN research. The data sets representing all the problems were real world data and obtained from the UCI machine learning benchmark repository (http://www.ics.uci.edu/~mlearn/MLRepository.htm). The characteristics of data sets are given in Table I.

### A. Data Set Description

The following subsections briefly describe the data set used in this study. The characteristics of the data sets are summarized in Table I. The detailed descriptions of the data sets are available at ics.uci.edu in directory /pub/machine-learning-databases [31], [32].

### A1. The breast cancer problem

The purpose of this problem is to diagnose a breast tumor as either benign or malignant based on cell descriptions gathered by microscopic examination. Input attributes are for instance the clump thickness, the uniformity of cell size and cell shape, the amount of marginal adhesion, and the frequency of bare nuclei. The data set representing this problem contained 699 examples. Each example consisted of nine-element real valued vectors. This is a two-class problem. All inputs are continuous; 65.5% of the examples are benign. This makes for entropy of 0.93 bits per example. This data set was created based on the "breast cancer Wisconsin" problem data set from the UCI repository of machine learning databases.

## A2. The diabetes problem

The objective of this problem is to diagnose whether a Pima Indian individual is diabetes positive or not based on his/her personal data, including age, number of times pregnant, and the results of medical examinations (e.g. blood pressure, body mass index, result of glucose tolerance test, etc.). There are 768 examples in the data set, each of which consisted of eight-element real valued vectors. This is a two-class problem. All inputs are continuous and 65.1% of the examples are diabetes negative; entropy 0.93 bits per example. This data set was created based on the "Pima Indians diabetes" problem data set from the UCI repository of machine learning databases.

## A3. The lenses problem

This problem uses a database for fitting contact lenses. The database is complete and noise free and contains 24 examples. These examples highly simplified the problem. The attributes do not fully describe all the factors affecting the decision as to which type, if any, to fit. All attributes are nominal. This is three-class problem: the patient should be fitted with hard contact lenses, soft contact lenses and no contact lenses.

**Table I** Characteristics of data sets.

| Data Sets | No. of Examples | Input Attributes | Output Classes |
|---|---|---|---|
| Breast Cancer | 699 | 9 | 2 |
| Diabetes | 768 | 8 | 2 |
| Lenses | 24 | 4 | 3 |

## B. Experimental Setup

In all experiments, one bias node with a fixed input 1 was used for the hidden and output layers. The learning rate was set between [0.1, 1.0] and the weights were initialized to random values between [-1.0, 1.0]. The number of training epochs $\tau$ was chosen between 5 and 20. Value of $\varepsilon$ for clustering was set between [0.1, 1.0]. Values of weight decay parameters $\varepsilon_1$, $\varepsilon_2$, were set between [0.05, .5] and [$10^{-4}$, $10^{-8}$] and $\beta$ was 10 for penalty function. A hyperbolic tangent function $\delta(y) = \frac{e^y - e^{-y}}{e^y + e^{-y}}$ was used as the hidden node activation function and a logistic sigmoid function $\sigma(y) = \frac{1}{1 + e^{-y}}$ as the output node activation function.

In this study, all data sets representing the problems were divided into two sets: the training set and the testing set. The numbers of examples in the training set and testing set were chosen to be the same as those in other works, in order to make the comparison with those works possible. The sizes of the training and testing sets used in this study are given as follows:

- Breast cancer problem: The first 350 examples are used for the training set and the rest 349 for the testing set.
- Diabetes problem: The first 384 examples are used for the training set and the rest 384 for the testing set.
- Lenses problem: The first 12 examples are used for the training set and the rest 12 for the testing set.

## C. Experimental Results

Tables II-IV show the ANN architectures produced by REANN and training epochs over 10 independent runs on three different classification problems. The initial architecture was selected before applying the constructive algorithm, which was used to determine the number of nodes in the hidden layer. The intermediate architecture was the outcome of the constructive algorithm, and the final architecture was the outcome of pruning algorithm used in REANN.

It is seen that REANN can automatically determine compact ANN architectures for all problems we consider in this work. For example, for the breast cancer data, the average number of nodes and connections were 6.8 and 5.8, respectively. For the diabetes data, the average number of nodes and connections were 12.5 and 19.4 respectively. It is seen that REANN produced compact and large architectures for cancer and diabetes problem. This is reasonable because cancer is the one of the easiest problem while diabetes is one of hardest problem in ANNs. It is natural to require compact architecture for solving easy problems and large architectures for hard problems.

**Table II** ANN architectures and training epochs for the **breast cancer** problem. The results were averaged over 10 independent runs.

|  | Initial Architecture | | Intermediate Architecture | | Final Architecture | | No. of Epoch |
|---|---|---|---|---|---|---|---|
|  | No. of Node | No. of Connection | No. of Node | No. of Connection | No. of Node | No. of Connection |  |
| Mean | 12 (9-1-2) | 11 | 12.7 | 18.1 | 6.8 | 5.8 | 233.2 |
| Min | 12 (9-1-2) | 11 | 12 | 11 | 5 | 5 | 222 |
| Max | 12 (9-1-2) | 11 | 14 | 33 | 10 | 9 | 245 |

**Table III** ANN architectures and training epochs for the **diabetes** problem. The results were averaged over 10 independent runs.

|  | Initial Architecture | | Intermediate Architecture | | Final Architecture | | No. of Epoch |
|---|---|---|---|---|---|---|---|
|  | No. of Node | No. of Connection | No. of Node | No. of Connection | No. of Node | No. of Connection |  |
| Mean | 11 (8-1-2) | 10 | 13.2 | 30 | 12.5 | 19.4 | 302.6 |
| Min | 11 (8-1-2) | 10 | 12 | 20 | 12 | 14 | 279 |
| Max | 11 (8-1-2) | 10 | 14 | 40 | 13 | 24 | 326 |

**Table IV** ANN architectures and training epochs for the **lenses** problem. The results were averaged over 10 independent runs.

|  | Initial Architecture | | Intermediate Architecture | | Final Architecture | | No. of Epoch |
|---|---|---|---|---|---|---|---|
|  | No. of Node | No. of Connection | No. of Node | No. of Connection | No. of Node | No. of Connection |  |
| Mean | 8 (4-1-3) | 7 | 9.1 | 14.7 | 8.9 | 12.1 | 109.2 |
| Min | 8 (4-1-3) | 7 | 8 | 7 | 8 | 7 | 97 |
| Max | 8 (4-1-3) | 7 | 10 | 21 | 10 | 17 | 128 |

Figs. 7-8 show the smallest of the pruned networks over 10 runs for breast cancer and diabetes problems. The pruned network for breast cancer problem has only 1 hidden node and 5 connections. The accuracy of this network was 96.275%. In this example, only three input attributes $A_1$, $A_6$ and $A_9$ were important and only three discrete values of hidden node activations were needed to maintain the accuracy of the network. The discrete values found by the heuristic clustering algorithm were 0.987, -0.986 and 0.004. The weight of the connection from the hidden node to the first output node was 3.0354 and to the second output node was –3.0354.

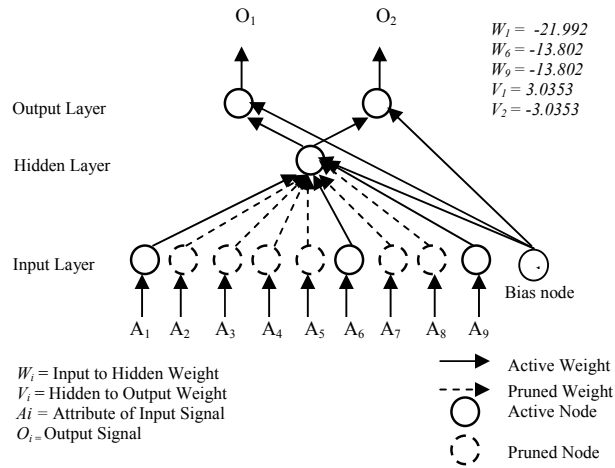

Fig. 7 A pruned network for breast cancer problem.

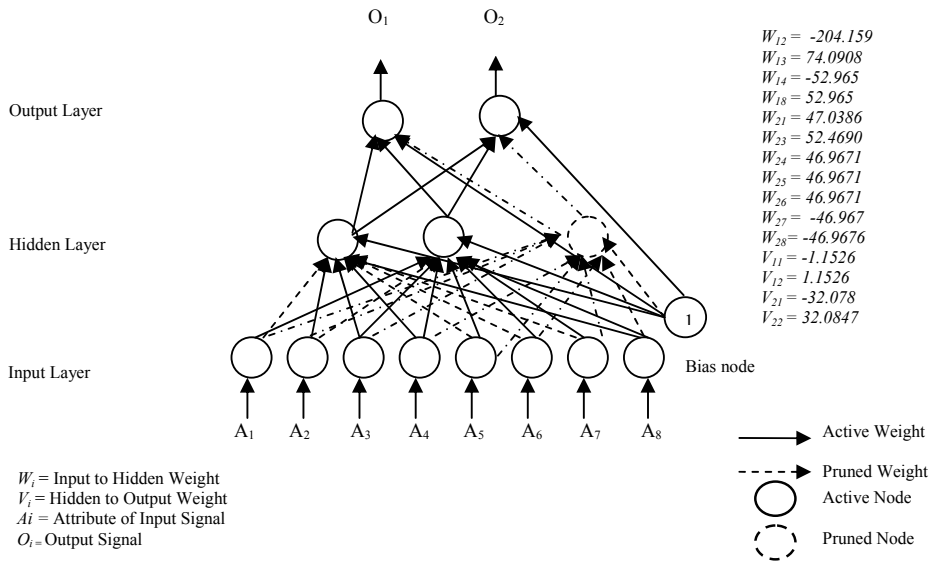

Fig. 8 A pruned network for diabetes problem.

The pruned network for diabetes problem has only 2 hidden nodes. No input nodes were pruned by pruning algorithm. One hidden node was pruned since all the connections to and from the node were pruned. The accuracy was 76.56 %. The weight of the connection from the first hidden node to the first output node was -1.153 and to the second output node was 1.153 and the weight of the connection from the second hidden node to the first output node was -32.078 and to the second output node was 32.084.

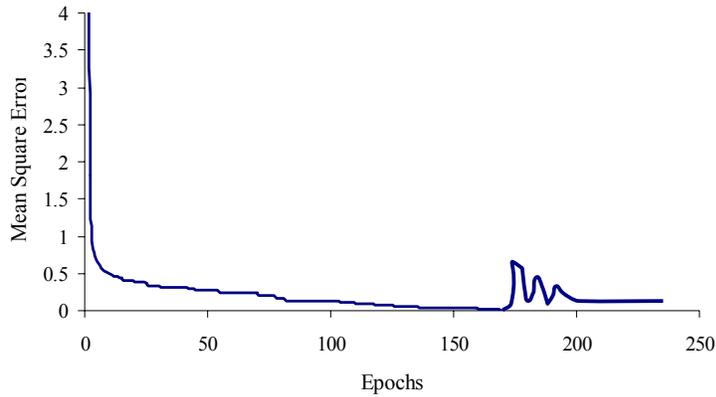

**Fig. 9** Training time error for breast cancer problem.

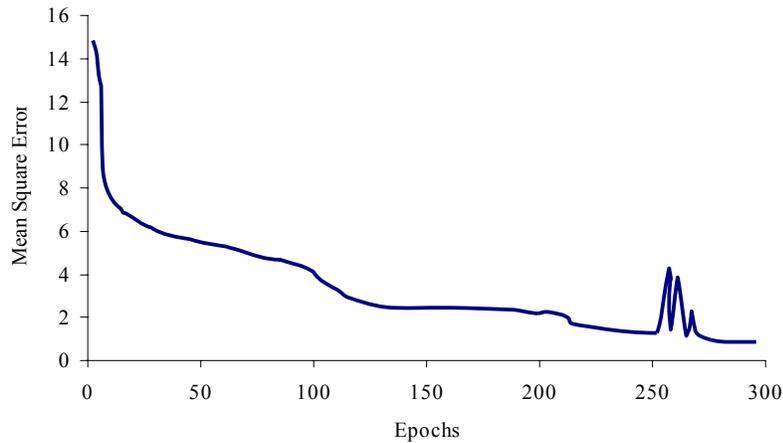

**Fig. 10** Training time error for diabetes data.

Figs. 9-10 show the training processes of REANN for breast cancer and diabetes problems. For breast cancer problem, it is observed that the training error decreases as training process progresses. After some training epochs, the training error is almost constant and then it is fluctuated for some epochs. The fluctuation is due to the pruning process of REANN. As the network was retrained after the pruning process, the network achieves the previous training error. The similar phenomena are also observed for the diabetes problem.

**Table V** Number of extracted rules and their classification accuracy for three different problems.

| Data Sets | No. of Extracted Rules | Rules Accuracy |
|---|---|---|
| Breast Cancer | 2 | 96.28 % |
| Diabetes | 2 | 76.56 % |
| Lenses | 8 | 100 % |

Table V shows the number of the extracted rules and their accuracy for three different problems. It is observed that two rules are sufficient to solve the breast cancer and diabetes problems. The accuracy was 100% for lenses classification, because the lower number of examples and the number of rules generated by REANN is 8.

## C1. Extracted rules

The number of rules extracted by REANN and the accuracy of the rules were described in Table V, but the visualization of the rules in terms of the original attributes ware not discussed. The aim of this subsection is to show what kinds of rules are generated by REANN for different problems. The number of conditions per rule and the number of rules extracted were also visualized here.

**The breast cancer problem**

> **Rule 1:** If Clump thickness ($A_1$) <= 0.6 and Bare nuclei ($A_6$) <= 0.5
>   and Mitosis($A_9$) <= 0.3, then benign
> **Default Rule:** malignant.

**The diabetes problem**

> **Rule 1:** If Plasma glucose concentration ($A_2$) <= 0.64
>   and Age ($A_8$) <= 0.69
>   then tested negative
> **Default Rule:** tested positive.

**The lenses problem**

> **Rule 1:** If Tear Production Rate ($A_4$) = reduce then no contact lenses
> **Rule 2:** If Age ($A_1$) = presbyopic and Spectacle Prescription ($A_2$) =
>   hypermetrope and Astigmatic ($A_3$) = yes then no contact lenses
> **Rule 3:** If Age ($A_1$) = presbyopic and Spectacle Prescription ($A_2$)
>   = myope and Astigmatic ($A_3$) = no then no contact lenses
> **Rule 4:** If Age ($A_1$) = pre-presbyopic and Spectacle Prescription ($A_2$) =
>   hypermetrope and Astigmatic ($A_3$) = yes and Tear Production Rate ($A_4$) =
>   normal then no contact lenses
> **Rule 5:** If Spectacle Prescription ($A_2$) = myope and Astigmatic ($A_3$) = yes
>   and Tear Production Rate ($A_4$) = normal then hard contact lenses
> **Rule 6:** If Age ($A_1$) = pre-presbyopic and Spectacle Prescription ($A_2$) =
>   myope and Astigmatic ($A_3$) = yes and Tear Production Rate ($A_4$)
>   = normal then hard contact lenses
> **Rule 7:** If Age ($A_1$) = young and Spectacle Prescription ($A_2$) = myope
>   and Astigmatic ($A_3$) = yes and Tear Production Rate ($A_4$)
>   = normal then hard contact lenses
> **Default Rule:** soft contact lenses.

## D. Comparisons

This section compares experimental results of REANN with the results of other works. A rule with many conditions is harder to understand than a rule with fewer conditions. Too many rules also hinder humans understanding of the data under examination. In addition to understandability, rules without generalization are not much of use. Hence, the comparison is performed along three dimensions: predictive accuracy, average number of conditions for a rule, and number of rules.

S. M. Kamruzzaman and Md. Monirul Islam
An Algorithm to Extract Rules from Artificial Neural Networks for Medical Diagnosis ProblemsThe primary aim of this work is not to evaluate REANN in order to gain a deeper understanding of rule extraction without an exhaustive comparison between REANN and all other works.

Table VI compares the REANN results of the breast cancer problem with those produced by NN RULES [10], DT RULES [10], C4.5 [29], NN-C4.5 [33], OC1 [33], and CART [34] algorithms. All algorithms achieve high predictive accuracy rates. However, REANN achieved the best performance and NN RULES was the closest second. It also outperformed other algorithms in terms of the number of rules. The number of rules extracted by REANN is 2 while it was 4 for NN RULES and 7 for DT RULES.

Table VII compares the REANN results of the diabetes problem with those produced by NN RULES, C4.5, NN-C4.5, OC1, and CART algorithms. REANN achieved 76.56% accuracy although NN-C4.5 was closest second with 76.4% accuracy. REANN is also outperformed all other algorithms in terms of average number of conditions in a rule and number of rules.

Table VIII compares the REANN results of the lenses data with those produced by PRISM [35]. Both algorithms achieved 100% accuracy because the lower number of examples. REANN extracted 8 rules, while PRISM extracted 9 rules.

**Table VI** Performance comparison of REANN with other algorithms for the **breast cancer** problem. '-' means not available.

| Data Set | Feature | REANN | NN RULES | DT RULES | C4.5 | NN-C4.5 | OC1 | CART |
|---|---|---|---|---|---|---|---|---|
| Breast Cancer | No. of Rules | 2 | 4 | 7 | - | - | - | - |
| | Avg. No. of Conditions | 3 | 3 | 1.75 | - | - | - | - |
| | Accuracy % | 96.28 | 96 | 95.5 | 95.3 | 96.1 | 94.99 | 94.71 |

**Table VII** Performance comparison of REANN with other algorithms for the **diabetes** problem. '-' means not available.

| Data Set | Feature | REANN | NN RULES | C4.5 | NN-C4.5 | OC1 | CART |
|---|---|---|---|---|---|---|---|
| Diabetes | No. of Rules | 2 | 4 | - | - | - | - |
| | Avg. No. of Conditions | 2 | 3 | - | - | - | - |
| | Accuracy % | 76.56 | 76.32 | 70.9 | 76.4 | 72.4 | 72.4 |

**Table VIII** Performance comparison of REANN with other algorithm for the **lenses** problem. '-' means not available.

| Data set | Feature | REANN | PRISM |
|---|---|---|---|
| Lenses | No. of Rules | 8 | 9 |
| | Avg. No. of Conditions | 3 | - |
| | Accuracy % | 100.0 | 100.0 |

## V. Conclusions

Although ANNs have been widely used to solve many problems, they are often viewed as black boxes. This work is an attempted to open up these black boxes by extracting rules from trained ANNs by the proposed rule extraction algorithm REANN. The experimental results on three different problems show that REANN can able to explain the functionality of ANN by extracting simple and concise rules. The predication accuracy of rules generated by REANN for different problems is also encouraging in comparison with exiting works.

56

REANN algorithm has some limitations that could be addressed in future work. REANN is not tested on classification problems having large number of output classes and regression problems. It would be interesting in the future to analyze REANN further on large classification and regression problems. The analysis would help to find the strength and weakness of REANN for such problems. In addition REANN is not considered the rule extraction technique for neuro-fuzzy network. A neuro-fuzzy network can be defined as a fuzzy system trained with some algorithm derived from ANNs. The integration of ANNs and fuzzy systems aims at the generation of a more robust, efficient and easily interpretable system where the advantages of each system are kept and their possible disadvantages are removed. Some ANN models such as the multilayer preceptron have been successfully applied to the training of neuro-fuzzy networks with back propagation algorithm to adjust the membership functions and connection weights of the processing nodes. In future, REANN could be applied for extracting rules for neuro-fuzzy network.

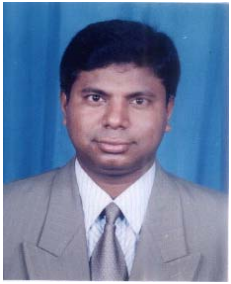

**S. M. Kamruzzaman** received the B. Sc. Engineering degree in Electrical and Electronic Engineering from the Bangladesh Institute of Technology (BIT), Dhaka, Bangladesh, in 1997, the M. Sc. Engineering degree in Computer Science and Engineering from Bangladesh University of Engineering and Technology (BUET), Dhaka, Bangladesh, in 2005. From 1998 to 2004, he was a Lecturer and Assistant Professor with the Department of Computer Science and Engineering, International Islamic University Chittagong (IIUC), Chittagong, Bangladesh.

In 2005, he moved to Manarat International University, Dhaka, Bangladesh as an Assistant Professor in the Department of Computer Science and Engineering. Currently he is working as an Assistant Professor in the Department of Information and Communication Engineering, University of Rajshahi, Bangladesh. His research interests include neural networks, communication engineering, data mining, Bangla language processing and pattern recognition.

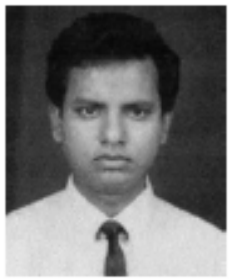

**Md. Monirul Islam** received the B. Sc. Engineering degree in Electrical and Electronic Engineering from the Bangladesh Institute of Technology (BIT), Khulna, Bangladesh, in 1989, the M. Sc. Engineering degree in Computer Science and Engineering from Bangladesh University of Engineering and Technology (BUET), Dhaka, Bangladesh, in 1996, and the Ph.D. degree in Evolutionary Robotics from Fukui University, Fukui, Japan, in 2002. From 1989 to 2002, he was a Lecturer and Assistant Professor with the Department of Electrical and Electronic Engineering, BIT, Khulna.

In 2003, he moved to BUET as an Assistant Professor in the Department of Computer Science and Engineering. Currently he is working as an Associate professor with the same department. His research interests include evolutionary robotics, evolutionary computation, neural networks, and pattern recognition.